# MSTDP: A More Biologically Plausible Learning

Shiyuan Li

## Abstract

Spike-timing dependent plasticity (STDP) which observed in the brain has proven to be important in biological learning. On the other hand, artificial neural networks use a different way to learn, such as Back-Propagation or Contrastive Hebbian Learning. In this work, we propose a new framework called mstdp that learn almost the same way biological learning use, it only uses STDP rules for supervised and unsupervised learning and don't need a global loss or other supervise information. The framework works like an auto-encoder by making each input neuron also an output neuron. It can make predictions or generate patterns in one model without additional configuration. We also brought a new iterative inference method using momentum to make the framework more efficient, which can be used in training and testing phases. Finally, we verified our framework on MNIST dataset for classification and generation task.

## Introduction

Almost every living creature has the ability to learn, but how the brain learns is still under studied by neuroscientists. Spike-timing dependent plasticity (STDP) [4, 13, 16] is believed to be a key fundamental of learning process in brains, and which can be described in very simple mathematical forms [1]. This has major impact on modern computational neuroscience because it's different from common machine learning algorithms, like Back-Propagation [2], Contrastive Hebbian Learning [5] or Simulated Annealing [18].

There are so many differences between biological learning and learning algorithms of artificial neural networks. Some properties that biological learning has but machine learning don't are: (i) the information needed for one neuron to perform learning algorithms is all in that neuron. It doesn't require other information like a global loss or post synapse neuron behavior. Neurons also don't need the information to tell them when to learn, like clamped phase or free phase in Contrastive Hebbian Learning (ii) There is no difference between the training process and the inference process. Animals don't distinguish training or testing, they are learning every moment and using what they have learned every moment. The training process just fed data into the model, and the model will learn on its own. (iii) Some observed biological characteristics, like asymmetric weights and temporal dynamics of neurons (when a neuron is activated, it first violently spikes and then drops to a value higher than the inactive state. Also, when a neuron is inactivated, it first strongly suppressed and then increases to a value below the activation state).

There are many machine Learning algorithms that can satisfy some of the properties above. The target propagation algorithm [12] computes local errors at each layer of the network using information about the target, which not need a global loss but still need the local post synapse neurons to propagated error signal. The recirculation algorithm [11] don't need any other information but the derivatives of the neurons, unfortunately it need symmetries weight. The Contrastive Hebbian learning (CHL) [5, 14, 15] which based on the Hebbian learning rule does not require knowledge of any derivatives. However, CHL have to tell the network to do different algorithms in clamped phase and free phase and it also requires synaptic symmetries and it need post synapse information. Others [17] use difference as a target to perform back-propagation. It has asymmetries weight, but the learning method is different from real neurons. All algorithms above have different training and inference processes, and they don't have the temporal dynamics.

In this paper, we proposed a new learning framework called Momentum Spike-timing dependent plasticity (mstdp), it satisfy all those properties above and has the same performance as existing machine learning algorithms. The mstdp method requires only the input of the neuron and derivative of the neuron to perform learning. There is no difference between the training process and inference process, and it doesn't need symmetric weights and has neural temporal dynamics.

Artificial neural networks commonly use different learning algorithms for supervised or unsupervised learning. While brain can also predict or generate, but doesn't distinguish between supervised or unsupervised learning. Animals learn their environment to survive and evolve, but they doesn't have a teacher or supervise information. Human babies may learn how to talk from their parents, but parents can't teach them how to see and hear. When a parent teaches a baby by pointing at a dog, they didn't teach the neurons in baby's brain the meaning of dog, they just linked the look of a dog and the sound of the word "dog" in the baby's brain. So we think there is no real supervise learning in biological learning, they just learn to link information and adapt to the environment, just like unsupervised learning algorithms learn to adapt to the data.

Motivated to create a more biologically plausible learning, we convert supervised learning problem into a unsupervised learning problem by making input neurons in the network also output neurons. Which is same to biological observations that photoreceptor neurons also has feedback connections. The model performs unsupervised learning like an auto-encoder by concatenate the input data with the labels. This automatically make prediction problem a in-painting problem by clamped the input data to get the labels. We can also generate samples from the model by clamped the label to get the input data.

The above modification will make it difficult for the model to converge. We solve this by add momentum in the iterative inference. Momentum Stochastic Gradient Descent is a well-known method for training Back-Propagation neural networks, It can effectively solve the problem of training jump into a local minimum. It can also be used for iterative inference processes to get better result. We use momentum iterative inference for both training and inference. We also smoothing the path in different state of the network to speed up the inference process.

The proposed learning framework can solve different problems, we verified it on MNIST dataset for classification and generation problems.

## Model and Methods

### 2.1 Model

We use the same model as [6], Suppose there are several neurons in the network, Every neuron has an internal state $s$, every two neurons are connected with a weight, and every neurons has a bias. Classical leaky integrator neural equation is used to calculate neurons behaviors. It follows:

$$\frac{ds_i}{dt} = \varepsilon(R_i(s) - s_i) \tag{1}$$

where $R_i(s)$ represents the pressure on neuron i from the rest of the network, while $\varepsilon$ is the time constant parameter. Moreover, suppose $R_i(s)$ is of this form:

$$R_i(s) = \rho'(s_i)(\sum_{j \neq i} W_{j,i} \rho(s_j) + b_i) \tag{2}$$

Where $W_{j,i}$ is the weight from the $j_{th}$ neuron to the $i_{th}$ neuron, $b_i$ is the bias of the $i_{th}$ neuron and $\rho$ is a nonlinear function. The purpose of this formula is to go down the energy function, which is defined by [6]:

$$E(\theta, s) = \frac{1}{2}\sum_i s_i^2 - \frac{1}{2}\sum_{i \neq j} W_{i,j} \rho(s_i)\rho(s_j) - \sum_i b_i \rho(s_i) \tag{3}$$

Where $\theta$ is the parameters in the model, in this case $W$ and $b$. Derive $E$ with respect to $s$ and with (1), we can get:

$$\frac{ds}{dt} = -\varepsilon \frac{\partial E}{\partial s}(\theta, s) \tag{4}$$

So the dynamics of the network is to perform Gradient Descent in $E$, each fixed-point for $s$ will correspond to a local minimum in $E$.

## 2.2 STDP rule

Spike-Timing Dependent Plasticity (STDP) is considered the main form of learning in brain, it relates the change in synaptic weights with the timing difference between spikes in post synaptic neurons and pre synaptic neurons. Experimental in [1] show that the stdp rule can also be form as:

$$\frac{dW_{i,j}}{dt} = \alpha \rho(s_i)\frac{ds_j}{dt} \tag{5}$$

Where α is the learning rate. In this paper, we change the form of stdp rule to:

$$\frac{\Delta W_{i,j}}{\Delta t} = \alpha \rho(s_i)\frac{\Delta \rho(s_j)}{\Delta t} \tag{6}$$

Which is more similar to the CHL rule since:

$$\rho_c(s_i)\rho_c(s_j) - \rho_f(s_i)\rho_f(s_j) = (\rho_f(s_i) + \Delta\rho(s_i))(\rho_f(s_j) + \Delta\rho(s_j)) - \rho_f(s_i)\rho_f(s_j)$$

$$= \rho_f(s_i)\Delta\rho(s_i) + \rho_f(s_j)\Delta\rho(s_j) + \Delta\rho(s_i)\Delta\rho(s_j)$$

Where $\rho_c$ is the clamped phase fixed-point and $\rho_f$ is the free phase fixed-point, we ignore the term $\Delta\rho(s_i)\Delta\rho(s_j)$ and get (6).

Different from CHL, we don't learn on phases. Every iteration will perform this learn rule, and the clamped phase and free phase are auto turn after a few iterations, don't synchronous to data. Every data have several clamped phases and free phases.

Therefore, the overall learning rule for $i_{th}$ neuron in this paper are:

$$\Delta W_{i,j} = \alpha \Delta \rho(s_i)\rho(s_j) \tag{7}$$

$$\Delta b_i = \alpha \Delta \rho(S_i)$$

## 2.3 Add inputs

There are some input neurons in the network, so we split **s** into two parts, **s_vis** and **s_hid**, **s = (s_vis, s_hid)**. For the input neurons **s_vis** we added another pressure to push it towards the input data:

$$\frac{ds_i}{dt} = \beta(data_i - s_i) \tag{8}$$

Where **data_i** is the input data for the $i_{th}$ input neuron. Which is equivalent to add another term in the energy function **E**:

$$E(\theta, s) = \frac{1}{2}\sum_i S_i^2 - \frac{1}{2}\sum_{i \neq j} W_{i,j} \rho(s_i)\rho(s_j) - \sum_i b_i \rho(s_i) + \frac{1}{2}\beta \sum_{i \in s_{vis}} (data_i - s_i)^2 \tag{9}$$

This is the same idea as in [7], the different between us is that **β** in this paper is not infinite for input neurons, and we don't distinguish input and output.

We define:

$$C(data, s) = \frac{1}{2}\beta \sum_{i \in s_{in}} (data_i - s_i)^2 \tag{10}$$

So the training process is to make **C** smaller.

## 2.4 Momentum inference

We use momentum to help iterative inference, so we change (1) into:

$$\frac{ds_i}{dt} = \varepsilon v_i \tag{11}$$

Where **v_i** is the velocity for **s_i**, and we update **v_i** use:

$$v_{i_t} = m(R_i(s) - s_i) + (1 - m)v_{i_{t-1}} \tag{12}$$

Where **m** is the inertia parameter.

## 2.5 Training

For training, we first initialize **W** and **b** randomly, and also randomly choose an **s** as initial.

Than we choose a data sample, set **β** to a positive value and do a few iteration, like the clamped phase in CHL. During the clamped phase we change **W**, **b** use (7) at every iteration. After that we set **β** to a 0 and do a few iteration, like the free phase in CHL. We repeat those clamped phase and free phase several times with the same data.

We loop this process for each data until convergence. The whole algorithms is demonstrate in algorithms 1.

---

Algorithms 1: Momentum Spike-timing Dependent Plasticity. **ε** is the iteration step for **s** to decrease **E**, **β** is the iteration step for **s** to decrease **C**, **α** is learning rate, **epochs** is training times, **T** is the times we do clamped phase and free phase for each data, **iteration** is iteration times in each relaxation.

---

Require: **ε, β, α, epochs, T, iteration**

Initialize **W,b** and **s** randomly

for n ← 1, . . . , epochs do

    data = data$_n$

    for t ← 1, . . . ,T do

        **β** = 1.0

        for i ← 1, . . . , iteration do

            Update **s** use (11)

            Update **W**, **b** use (7)

        end for

        **β** = 0.0

        for i ← 1, . . . , iteration do

            Update **s** use (1)

        end for

    end for

end for

---

Notice that we don't reinitialize **s** after data changed, so the first clamped phase and free phase will change **s** from one fixed-point to a different fixed-point. This moving path will also be learned by the stdp rule.

The frequency neurons change clamped phase into free phase doesn't related to the frequency we change data, just like in biologically learning process we can't teach the neurons when to learn in clamped phase or free phase.

## 2.6 Testing

The test process is same to training process, except the learning rate is zeros and we can do fewer clamped and free phases.

## 2.7 Smoothing derivative

We also smoothing the path between different fixed-points by make derivative of every point on the path smaller, in order to make the network easier to jump from one fixed-point to another fixed-point. We perform this by calculate the derivative of **ds** to **W**, adjust **W** to make **ds** smaller. **ds** is:

$$ds_i = \rho'(s_i)(\sum_{j \neq i} W_{j,i}\rho(s_j) + b_i) - s \qquad (13)$$

So the derivative of **ds** to **W**, **b** is:

$$\frac{\partial^2 E}{\partial s \partial W_{i,j}} = \rho'(s_i)\rho(s_j) \qquad (14)$$

$$\frac{\partial^2 E}{\partial s \partial b_i} = \rho'(s_i) \qquad (15)$$

And the learning rule will be:

$$\Delta W_{i,j} = \rho'(s_i)\rho(s_j)$$

(16)

$$\Delta b_i = \rho'(s_i)$$

(17)

This will make the path between different fixed-points flatter and easier to jump.

## 2.8 Why this work

In the beginning, the model had some fixed-point and they didn't make any sense. By clamped the input neurons to data, we end up with a fixed-point that has low **C** to the current data. In each iteration, the learning rule will make the output of the neurons more like the next iteration's output, so the whole learning process will make the network easier to jump from the old fixed-point into the new fixed-point. When the jump is from a free fixed-point into the clamped fixed-point, it will make the free fixed-point has lower **C**. When the jump is from one data's fixed-point into another (for example, from data$_i$'s fixed-point to data$_j$'s fixed-point), the opposite jump will cancellation the learning (from the data$_j$'s fixed-point into data$_i$'s fixed-point). So, from all, we make every data's free fixed-point has lower **C**.

Therefore, with those free fixed-points, the network can get some meaningful output without input. Just like humans can imagine things with their eyes closed, which is a property not find in other methods [5, 11, 12, 14, 15].

It has been proved that random feedback can play the same role as back-propagation [8,9]. So after training these input neurons will have the same value for the input data, and **C** will become very small.

The training process will create many fixed-points. When data changed, the new pressure of **C** drives **s** to leave the old fixed-point, and the pressure of other neurons will stop it from leave the current fixed-point. The force of **C** is usually not big enough to make **s** to jump out of current fixed-point, that's why we need the momentum to help it jump. The training process will pick the fixed-points that is easy to jump and make it have a lower **C**.

# Result

In this section, we will verify our algorithm on classification tasks and generation tasks. The dataset we use is the MNIST Handwritten Digit and Letter Classification dataset. The neural network has 784 + 10 input neurons for data and labels and the number of hidden neurons is 2048. Each input neuron is connected to all hidden neurons, and all hidden neurons are interconnected, but input neurons don't connected to each other.

We train our model on 10000 numbers for 500000 times and test on 500 numbers. We don't train with batches, which makes training and testing the same. The learning rate we use is 0.001 and there are 10 clamped and free phases for each data. For champed phases, **ε** is 0.2, **β** is 0.8 and **m** is 0.4. For free phases, **ε** is 0.2, **β** is 0 and **m** is 0. We use sigmoid4 as activate function to speed up the training process, which is slightly different form the sigmoid function (sigmoid4(x) = sigmoid(4x)). Parameters **W** and **b** are initialized with a uniform distributions **U**(−0.1, 0.1). For each clamped and free phase, we iterate 32 times to get **s**.

### 3.1 Classification

We clamped the input data and labels in training and only clamped input data and do three clamped phase and free phase to get the labels in testing. After training, it has an accuracy of 100% on the training set and 96% on the testing set.

Figure 1 shows the free phase fixed-point, 1st clamped phase fixed-point, 2nd clamped phase fixed-point and labels in a training loop. The free phase fixed-point is meaningless, but after jumping it will end up in a fixed-point pretty close to the label.

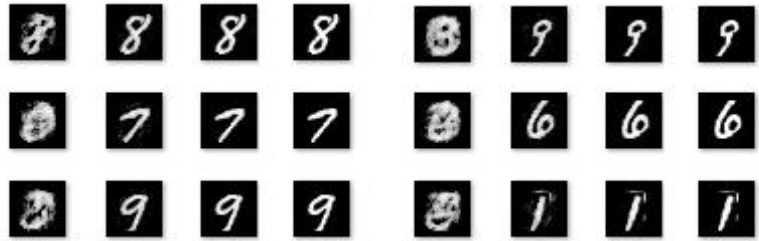

**Figure 1 there are 6 groups of data in the figure, for each group of data, from left to right: (a) random fixed-point. (b) 1st fixed-point after the clamped data. (c) 2nd fixed-point after clamped data. (d) labels**

The biological neurons have temporal dynamics, and momentum inference can lead to the same result. Figure 2 shows the similarity between momentum inference and real neurons. Left is the simulation in our paper, right [10] is the response of a real biological neuron to certain features. X axis represents time and Y axis represents intensity.

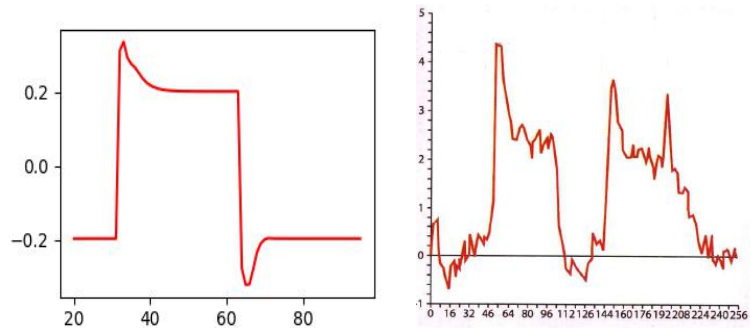

**Figure 2 left: simulations by the model. right: response of a real neurons.**

### 3.2 Generation

We can generate samples by randomly pick a fixed-point, but this usually get meaningless samples (figure 1 a). So we use conditional generation method to generate numbers by first randomly pick *s* and then do a few clamped and free phases with label clamped. The label will constrain the distribution of hidden neurons to get more reasonable result, so the network will jump to a number looks image in that label. Figure 3 shows some samples generated by our model.

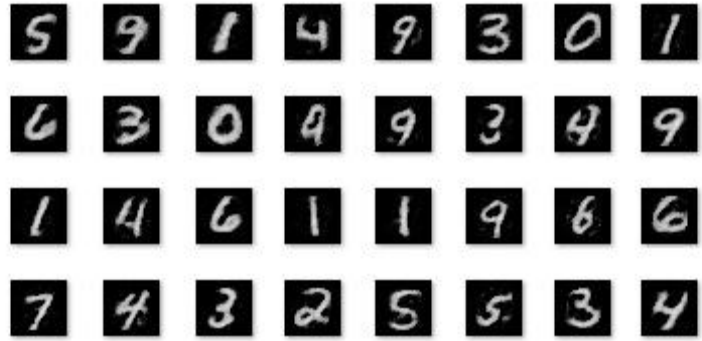

**Figure 3 some numbers generated by our model**

## Discussion

In this work, we introduced a framework for supervised learning and unsupervised learning. The way it learns is very similar to biological learning. But there are still some differences between the framework and real neurons: (i) real neurons has sparse representations, but our framework doesn't. (ii) in our framework, the network will finally stay at a fixed-point and no longer jump. But brain never stop thinking even without inputs. Instead, focus on one thing is difficult.

For (i), real neurons consume energy when they spike, and energy is precious to wild animals. Sparse representation make fewer neurons spike, which will help them save energy. So, whether sparse representation is help for better representation or just help for save energy still needs discussion.

For (ii), stay at one fixed-point doesn't help animals for rapid response to dangers in environment, so there must be some mechanism to help it jump out a fixed-point if stays in it too long. Real neurons will become tired when they spike. This may lead to a reduction in spikes and make it hard to stay in a fixed-point.

On the other hand, the math in (4) has some approximations, the derivative of *E* is actually:

$$\frac{dE}{ds_i} = \rho'(s_i)(\sum_{j \neq i} \frac{1}{2}(W_{j,i} + W_{i,j})\rho(s_j) + b_i)$$

We use *$W_{ij}$* instead of *($W_{ij}$+$W_{ji}$)/2* to simplify the calculation. If we consider *s* as a force, the space of *s* will become a force field. When the curl of the field is not zero, particles in the force field may move forever. The curl of the field is always zero when we use *($W_{ij}$+$W_{ji}$)/2*, but not always zero when use *$W_{ij}$*. This may also cause the network not stay.

The framework works like an auto-encoder network, which usually have a bottleneck. If an auto-encoder doesn't have a bottleneck, the network will tend to be trained into an identity map for every data. But bottlenecks also reduce the representation ability of the network. Our model may be a solution of this since the number of hidden neurons of our model can be greater than the number of dimensions of the input data.

Generating number images by randomly pick a fixed-point will get meaningless samples. This is due to the superfluous fixed-point that don't match any data. However this will make bad influence to the performance of the network. How to avoid this will further study in the future.

# References

[1] Y Bengio, T Mesnard, A Fischer, S Zhang, Y Wu, STDP as presynaptic activity times rate of change of postsynaptic activity. arXiv preprint arXiv:1509.05936.

[2] Y LeCun, BE Boser, JS Denker, Handwritten digit recognition with a back-propagation network, Advances in Neural Information Processing Systems, 1990

[3] JR Movellan, Contrastive Hebbian Learning in the Continuous Hopfield Model, Connectionist models, 1991

[4] G Bi, M Poo, Synaptic modifications in cultured hippocampal neurons: dependence on spike timing, synaptic strength, and postsynaptic neurons type. Journal of neuroscience, 18(24):10464-10472, 1998.

[5] P Baldi, F Pineda, Contrastive learning and neural oscillations. Neural Computation, 3(4):526–545, 1991.

[6] Y Bengio, A Fischer, Early inference in energy-based models approximates back-propagation. arXiv preprint arXiv:1510.02777.

[7] B Scellier, Y Bengio, Towards a Biologically Plausible Backprop, arXiv:1602.05179, 2016

[8] TP Lillicrap, D Cownden, DB Tweed, Random feedback weights support learning in deep neural networks, arXiv preprint arXiv:1411.0247

[9] G Detorakis, T Bartley, E Neftci, Contrastive Hebbian Learning with Random Feedback Weights, Neural Networks, 2019

[10] M Gazzaniga, R Ivry, G Mangun, Cognitive Neuroscience The Biology of the Mind, page 162.

[11] G Hinton, J McClelland, Learning representations by recirculation. In Neural information processing systems, pages 358–366, 1988.

[12] D Lee, S Zhang, A Fischer, Y Bengio. Difference target propagation. In Joint european conference on machine learning and knowledge discovery in databases, pages 498–515. Springer, 2015.

[13] H Markram, J L¨ubke, M Frotscher, B Sakmann. Regulation of synaptic efficacy by coincidence of postsynaptic aps and epsps. Science, 275(5297):213–215, 1997.

[14] JR Movellan. Contrastive hebbian learning in the continuous hopfield model. In Connectionist models: Proceedings of the 1990 summer school, pages 10 – 17, 1991.

[15] X Xie, H Seung. Equivalence of backpropagation and contrastive hebbian learning in a layered network. Neural computation, 15(2):441 – 454, 2003.

[16]L Zhang, W Tao, C Holt, W Harris, M Poo. A critical window for cooperation and competition among developing retinotectal synapses. Nature, 395(6697):37, 1998.

[17] Y Bengio, DH Lee, J Bornschein, T Mesnard, Towards biologically plausible deep learning, arXiv preprint arXiv:1502.04156

[18] E Aarts, J Korst, Simulated annealing and Boltzmann machines, 1988